\documentclass[10pt,twocolumn,letterpaper,dvipsnames]{article}
\usepackage[most]{tcolorbox}

\usepackage[pagenumbers]{cvpr}      %

\definecolor{cvprblue}{rgb}{0.21,0.49,0.74}
\usepackage[pagebackref,breaklinks,colorlinks,citecolor=cvprblue]{hyperref}
\usepackage{epigraph} 
\usepackage[utf8]{inputenc}
\usepackage{csquotes}
\usepackage{wrapfig}

\usepackage{graphicx}
\usepackage{xspace}
\usepackage{booktabs}

\definecolor{MyDarkGreen}{rgb}{0.02,0.6,0.02}

\newcommand{\Task}{Neural Object Kinematics\xspace}
\newcommand{\btask}{\textbf{Neur}al \textbf{O}bject \textbf{K}inematics\xspace}
\newcommand{\stask}{\textsc{NeuROK}\xspace}

\newcommand{\qq}{\mathbf{q}}

\newcommand{\xx}{\mathbf{x}}
\newcommand{\zz}{\mathbf{z}}
\newcommand{\ee}{\mathbf{e}}

\newcommand{\zdotdot}{\ddot{\mathbf{z}}}
\newcommand{\zdot}{\dot{\mathbf{z}}}

\newcommand{\rset}{\mathbb{R}}

\newcommand{\meshz}{{\mathcal{M}_{0}}}
\newcommand{\deformf}{{\phi}}
\newcommand{\econd}{{\mathcal{E}_{\text{cond}}}}
\newcommand{\evae}{{\mathcal{E}_{\text{VAE}}}}
\newcommand{\decoder}{{\mathcal{D}}}
\newcommand{\nsample}{{n_{\text{sample}}}}

\newcommand{\pfeat}{\mathbf{F}_\text{cond}}
\newcommand{\pfeatvae}{\mathbf{F}_\text{VAE}}
\newcommand{\fdim}{{F_\text{pos}}}
\newcommand{\ntokenscond}{{K}}
\newcommand{\ntokens}{{K}}
\newcommand{\dimtoken}{F_{\text{token}}}
\newcommand{\codebookcond}{\{\ee_i\}_{i=1}^{\ntokenscond}}
\newcommand{\dodt}{\frac{\mathrm{d}}{\mathrm{d}t}}

\makeatletter
\DeclareRobustCommand\onedot{\futurelet\@let@token\@onedot}
\def\@onedot{\ifx\@let@token.\else.\fi\xspace}
\makeatother

\newcommand{\supp}{Supp. Mat\onedot}

\usepackage{multirow}
\usepackage{makecell}

\usepackage[most]{tcolorbox}
\tcbuselibrary{skins, breakable}
\usepackage{amsmath} %

\tcbset {
  base/.style={
    arc=0mm, 
    bottomtitle=0.5mm,
    boxrule=0mm,
    colbacktitle=black!10!white, 
    coltitle=black, 
    fonttitle=\bfseries, 
    left=2.3mm,
    leftrule=1mm,
    right=3mm,
    title={#1},
    toptitle=0.75mm, 
  }
}

\tcbset{
  meshstep/.style={
    enhanced,
    colback=gray!6,
    colframe=gray!35,
    boxrule=0.5pt,
    arc=2pt,
    left=5pt,right=5pt,top=6pt,bottom=6pt,
    attach boxed title to top left={
      xshift=0.5em,
      yshift*=-1mm
    },
    boxed title style={
      enhanced,
      colback=gray!30!white,
      colframe=gray!50,
      arc=1.5pt,
      top=3pt,bottom=3pt,
      fontupper=\bfseries\footnotesize,
    },
    drop shadow={opacity=0.18},
  }
}

\tcbset{
  llmprompt/.style={
    enhanced,
    breakable,
    colback=gray!8,
    colframe=gray!40,
    boxrule=0.4pt,
    arc=2pt,
    left=6pt,right=6pt,top=6pt,bottom=6pt,
    width=\linewidth,
  }
}

\newtcbtheorem[auto counter]{definition}{Definition}{
                lower separated=false,
                colback=blue!3!white,
colframe=blue!40!black, 
base={#1}
  breakable
}{def}

\newtcbtheorem[auto counter]{theorem}{Theorem}{
    lower separated=false,
    colback=yellow!10!white,   %
    colframe=yellow!60!black,  %
    base={#1},
    breakable
}{thm}

\newtcbtheorem[auto counter]{example}{Example}{
                lower separated=false,
                colback=teal!3!white,
colframe=teal!50!black, 
base={#1},
  breakable
}{example}

\newcommand{\myparagraph}[1]{\vspace{0.5em}\noindent\textbf{#1}\xspace~}

\usepackage[capitalize]{cleveref}
\crefname{section}{Sec.}{Secs.}
\crefname{table}{Tab.}{Tables}
\crefname{figure}{Fig.}{Figures}
\crefname{tcb@cnt@example}{Ex.}{Examples}
\crefname{tcb@cnt@definition}{Def.}{Definitions}
\crefname{tcb@cnt@theorem}{Thm.}{Theorems}

\usepackage{longtable}
\usepackage{tabularx}
\usepackage{algorithm2e}

\def\paperID{*****} %
\def\confName{CVPR}
\def\confYear{2026}

\title{\textsc{NeuROK}: Generative 4D Neural Object Kinematics}

\author{~~~Chen Geng$^{1,*}$
\qquad
Guangzhao He$^{3,*}$
\qquad
Yue Gao$^{1,*}$
\qquad
\\
Yunzhi Zhang$^{1}$
\qquad
Shangzhe Wu$^{2}$
\qquad
Jiajun Wu$^{1}$ \\[0.8em]
$^1$Stanford University \qquad
$^2$University of Cambridge \qquad
$^3$Cornell University
}

\begin{document}

\twocolumn[
    \maketitle
    \begin{center}
    \vspace*{-25pt}
    \includegraphics[width=\textwidth]{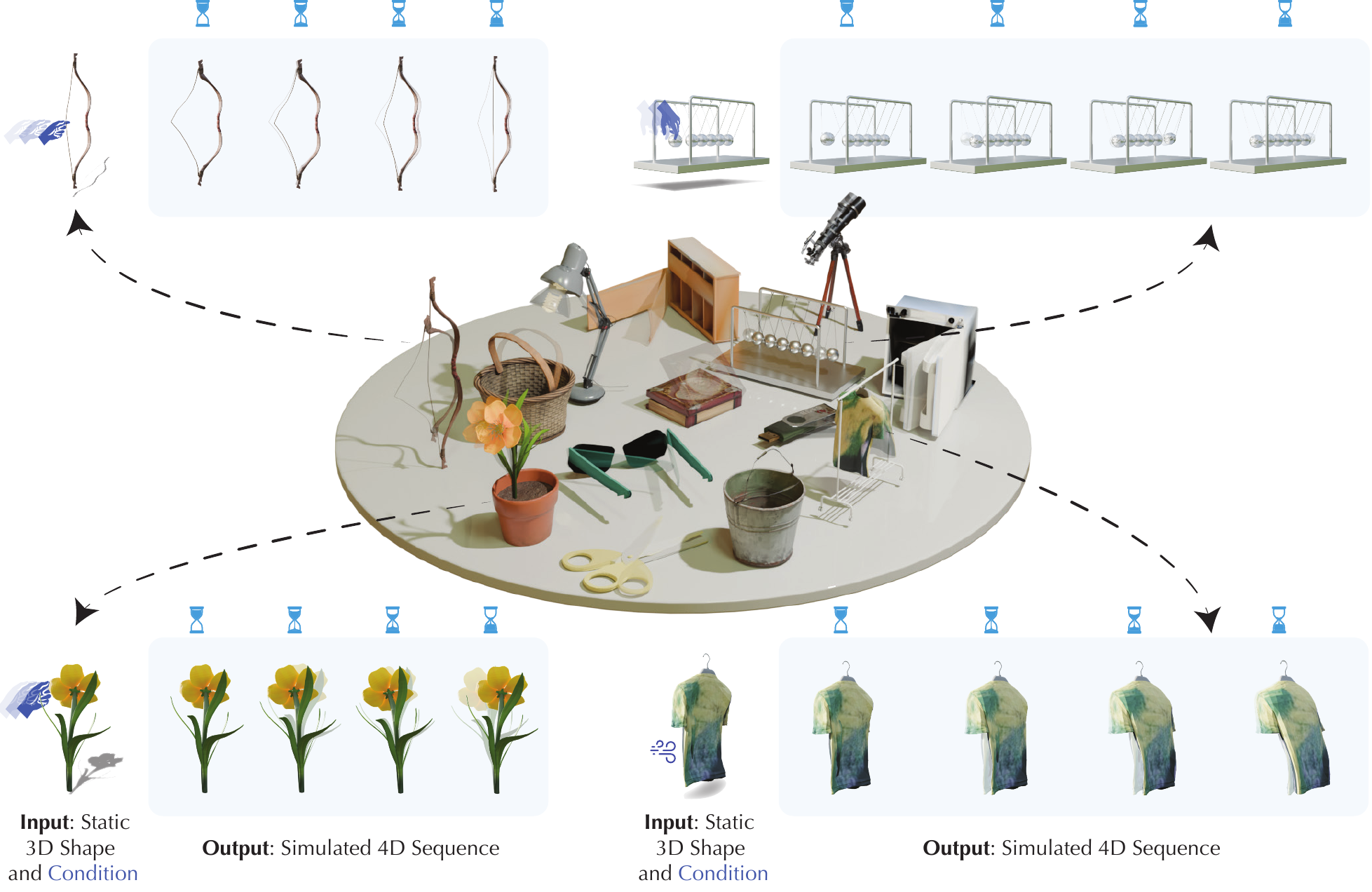}
    \vspace{-20pt}
    \captionof{figure}{ We present a versatile and scalable framework for generating simulative 4D dynamics of static 3D objects under physical conditions (\textit{e.g.}, forces, actions, velocities). Trained on a large-scale 4D shape dataset without any explicit physical annotations, our method does not rely on any inductive bias of the object's dynamic structure and therefore can be applied to various types of dynamic objects, ranging from elastic bodies, cloth, and continuum bodies, to multi-body objects. Project page: \url{https://chen-geng.com/neurok}
    }
    
    \label{fig:teaser}
\end{center}

    \bigbreak
]

\newcommand{\episize}{\fontsize{8.6pt}{10pt}\selectfont}

\begin{abstract}
\vspace{-5pt}
Data-driven approaches have revolutionized 3D vision, enabling transformers to effectively reconstruct and generate static 3D objects. However, generating simulative 4D dynamics---realistic temporal deformations of static objects under various physical conditions---remains challenging and often ad hoc, despite its importance in building comprehensive 3D world models. Most existing methods assume a predefined physical model and use system identification to estimate parameters, restricting these methods to specific categories and small-scale datasets. 

We propose that these restrictions can be overcome by learning a data-driven kinematic state parameterization for object-centric physical systems. Specifically, we learn both a latent space representing all possible states of the object and a decoder that maps any sampled latent to a plausibly deformed shape of the object. 
We refer to this parameterization as \btask (\stask), and learn a transformer-based encoder-decoder model on a curated large-scale 4D dataset. This formulation and the learned model significantly simplify the generation of simulative dynamics since we only need to consider the dynamics within a low-dimensional latent space from the Lagrangian mechanics' perspective in classical physics. 
We demonstrate the effectiveness and generality of this neural simulation framework across diverse dynamic object types, showing clear advantages over prior works. 

\end{abstract}

\begin{figure*}[t]
    \centering
    \vspace{-20pt}
    \includegraphics[width=1.\linewidth]{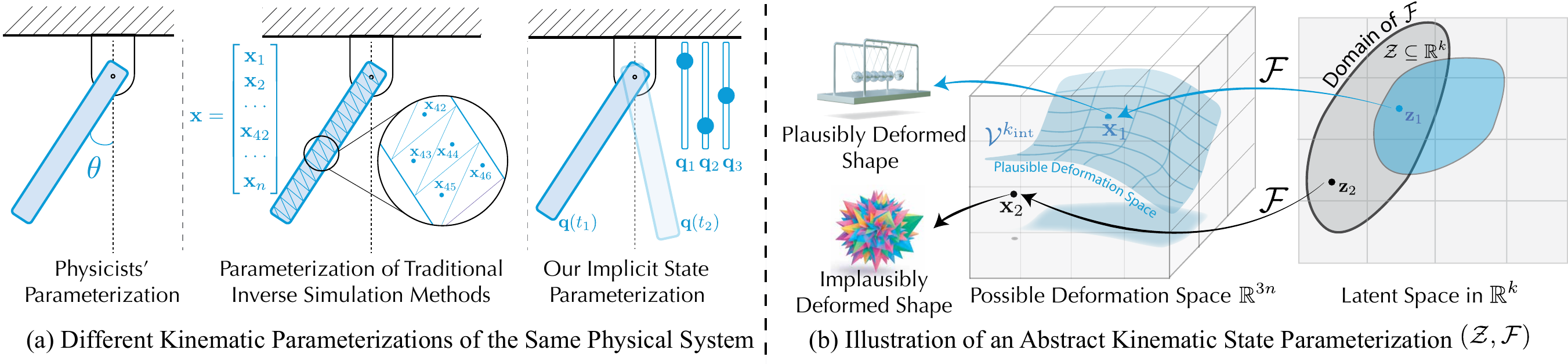}
    \vspace{-20pt}
    \caption{ \textbf{Kinematic state parameterization.} \textbf{(a)} Several kinematic state parameterizations can be used to describe a physical system. The symbolic parameterizations used in classical mechanics are concise yet not accessible in inverse problems. Traditional inverse simulation approaches use geometry-derived parameterizations, yet require dense physical constraints to solve the over-parameterized system. We instead learn low-dimensional parameterizations that are both compact and learnable from data. \textbf{(b)} As formally defined in \cref{def:kin_state_param}, a kinematic state parameterization studied in this paper is a pair $(\mathcal{Z}, \mathcal{F})$ which contains a latent manifold $\mathcal{Z}$ and a decoder $\mathcal{F}$ that maps a sampled latent to a vertex configuration. This definition explicitly includes those kinematic state parameterizations that are not compact.
    \vspace{-10pt}
    \label{fig:kin_state_param}
    }
\end{figure*}

\vspace{-6pt}
\section{Introduction}

\epigraph{\episize{These quantities need not be the Cartesian co-ordinates of the particles, and the conditions of the problem may render some other choice of coordinates more convenient.} }{--- \episize{\textit{L. Landau \& E. Lifshitz, in Mechanics, 1960}}}

Given a 3D geometric snapshot of a dynamic object, humans can intuitively imagine how the object would react under different physical conditions, even without precise  knowledge of the governing physical equations. However, in the community of generative AI, generating such 4D reactive behaviors with no reliance on any category-specific physical priors is far from trivial, despite the importance of this capability in constructing 3D world models for embodied AI or robotics~\cite{mandi2025dexmachina,li2023behavior}.

A long-standing view holds that generating such 4D simulative dynamics demands a comprehensive physical understanding of the object. This is epitomized by most existing works~\cite{zhang2024physdreamer,xie2024physgaussian} that generate 4D dynamics by adopting predefined category-specific physical models and estimating their parameters with system identification. While this paradigm is effective for target object categories (\eg, articulated objects, continuum bodies, and cloth), it struggles to generalize beyond these predefined categories, and, more importantly, offers limited scalability to large-scale 4D datasets comprising diverse dynamic structures.

Is it possible to build a general-purpose simulator that generates such 4D motions without any category-specific inductive bias? We argue that this is achievable by re-considering a critical yet long-overlooked piece: the \textbf{kinematic state parameterization} of dynamic objects. As illustrated in \cref{fig:kin_state_param}, a kinematic state parameterization defines the configuration space of state vectors that fully specify an object's geometry. Most existing approaches~\cite{zhang2024physdreamer,lin2025omniphysgs,xie2024physgaussian} adopt a kinematic state parameterization naturally inherited from the object's shape representation, \eg, a dense particle set derived from mesh discretizations. While effective, this choice leads to an over-parameterized system, and thus necessitates category-specific physical constraints to prevent the system from being under-determined.

We revisit this important factor by introducing an automatically-discovered kinematic state parameterization scheme --- \btask (\stask) --- a latent space from which any vector sampled can be decoded into a plausible deformation of the modeled object. With this learned parameterization, the physical system can be greatly simplified: we only need to model the transition between low-dimensional latent vectors, similar to how a pendulum system can be simplified through a symbolic parameterization (\cref{fig:kin_state_param}(a)). This data-driven parameterization leads to a universal framework that simulates system dynamics from the perspective of Lagrangian mechanics~\cite{landau1960mechanics}, where category-agnostic energy functions are defined over the latent states and dynamics are directly derived using Euler-Lagrange equations.

This framework forms a versatile and scalable pipeline for generative simulation of dynamic objects. Its core learning component, \stask, adopts a transformer-based~\cite{vaswani2017attention} encoder-decoder architecture that learns to encode a static 3D object into a latent distribution over its possible kinematic states and to decode any sampled latent vector into a corresponding deformation field. The model can be trained solely on 4D geometric trajectories of 3D objects, eliminating any need for physical or action annotations. Moreover, this framework relies on a minimal inductive bias --- that  the object's deformation space is low-dimensional --- making it broadly applicable to diverse dynamic objects.

We validate our framework by curating a large-scale 4D object dataset, training a feed-forward \stask model, and generating 4D dynamics across a wide range of objects. We evaluate its performance by comparing against existing methods, demonstrating its superior generalizability and effectiveness. To the best of our knowledge, this is the first data-driven framework capable of simulating object-centric physical systems without any reliance on heuristic priors or physical annotations.

\section{Related Work}
\label{sec:related}
\vspace{-5pt}

\myparagraph{Physically-Inspired 4D Generation.} Existing approaches to generating 4D simulative dynamics typically follow a two-step paradigm: finding a physical model of the targeted domain, and determining its parameters with system identification. This includes directly modeling physical properties of rigid objects~\cite{liu2024physgen,xu2019densephysnet,zhai2024physical}; modeling elastic objects with MPM~\cite{li2022plasticitynet,jiang2016material,zhang2024physdreamer,xie2024physgaussian,li2023pac,chen2025physgen3d,li2025wonderplay,lin2025omniphysgs,cai2024gic,lin2024phys4dgen,lin2025visionlaw,kaneko2024improving,dagli2025vomp,zhao2025physsplat,mittal2025uniphy}, projective dynamics~\cite{qiao2022neuphysics,du2021diffpd,bouaziz2023projective}, or geometry-agnostic elastic simulation methods~\cite{modi2024simplicits,chen2025vid2sim,feng2024pie}; using spring-mass to model deformable objects~\cite{jiang2025phystwin,zhong2024reconstruction}; predicting articulations to model articulated objects~\cite{xia2025drawer,chen2024urdformer,partnet_mobility,le2024articulate,lei2023nap,liu2024singapo,mandi2024real2code,noguchi2022watch,jiang2022ditto,kerr2024robot,liu2023paris,xu2025gaussianproperty}; and building physical models for cloth~\cite{li2022diffcloth,li2024diffavatar,macklin2016xpbd,guo2025pgc,zheng2024physavatar,li2025dress,liang2019differentiable}. While they perform well within specific domains, none can generate 4D motions without assuming a predefined dynamic structure. Our framework removes such structural biases, enabling general 4D simulative dynamics generation.

\myparagraph{Reduced-Order Simulation.}
Model reduction is a common technique in forward computer graphics, yet the focus is \emph{efficiency} rather than versatility. The goal of such approaches~\cite{viswanath2024reduced,lee2020model,chang2023licrom,chen2022crom,zong2023neural,sharp2023data,james2006precomputed,james2002dyrt,benner2015survey,fulton2019latent,shen2021high,romero2021learning,li2025self,wang2024neural,modi2024simplicits,chen2023model} is typically to accelerate an existing physical simulation system where all physical constraints are known, in contrast to our category-agnostic setting. These approaches typically train instance-specific neural networks to represent the reduced-order kinematic space for a specific object, rather than learning a generalizable, amortized-inference model on a large dataset as in our framework.

\myparagraph{Machine Learning for Dynamic Systems.} Beyond 3D vision, machine learning has also been used to model non-visual dynamic systems, typically through either physics-agnostic or physics-aware approaches. Physics-agnostic methods ~\cite{li2018learning,pfaff2020learning,sanchez2020learning,battaglia2016interaction,chang2016compositional,li20223d,li2024deformnet,shen2024action,schenck2018spnets} learn dynamics end-to-end --- often via GNNs --- using synthetic datasets of action-state pairs. Although effective in controlled settings, they struggle to generalize to real-world objects due to the scarcity of action-labeled data. In contrast, our method relies solely on 4D geometric supervision, offering greater scalability for graphics and 3D vision applications. Physics-aware methods assume known physical models and use neural networks to solve PDEs~\cite{raissi2019physics,li2020fourier,li2023learning,chen2023implicit}, learn constitutive laws~\cite{ma2023learning,mittal2025uniphy}, and learn discretization schemes~\cite{bar2019learning}. While demonstrating potential in producing accurate solutions, these approaches are unsuitable for our setting which makes no assumptions about dynamic structures. Closer to our formulation are methods that use neural networks~\cite{lutter2019deep,cranmer2020lagrangian,finzi2020simplifying,bhattoo2022learning} to model systems within the Lagrangian mechanics framework, but their focus is learning the system's Lagrangian from synthetic data rather than learning data-driven kinematic state parameterizations.

\myparagraph{Neural Deformation Priors.} Several graphics systems have also explored learning data-driven priors over object deformations, but most are category-specific~(\eg, for humans~\cite{loper2023smpl,pavlakos2019expressive}, faces~\cite{blanz2003face,booth20163d,bailey2020fast}, and animals~\cite{Zuffi:CVPR:2017}) and target other tasks --- most dominantly, for animating characters~\cite{huang2021arapreg,yang2023geolatent,maesumi2023explorable,he2025category,yang2023gencorres,jakab2021keypointdeformer,yoo2024neural,bozic2021neural,giebenhain2023learning,palafox2021npms,palafox2022spams,wang20193dn,yifan2020neural,zhou2020unsupervised} and controlling embodied agents~\cite{li2025controlling,liu2024differentiable,ruan2025prof}. We instead formalize this idea through the concept of kinematic state parameterization and demonstrate its huge potential as a general interface in physically-inspired 4D generation.

\begin{figure*}
    \centering
    \vspace{-20pt}
    \includegraphics[width=0.9\linewidth]{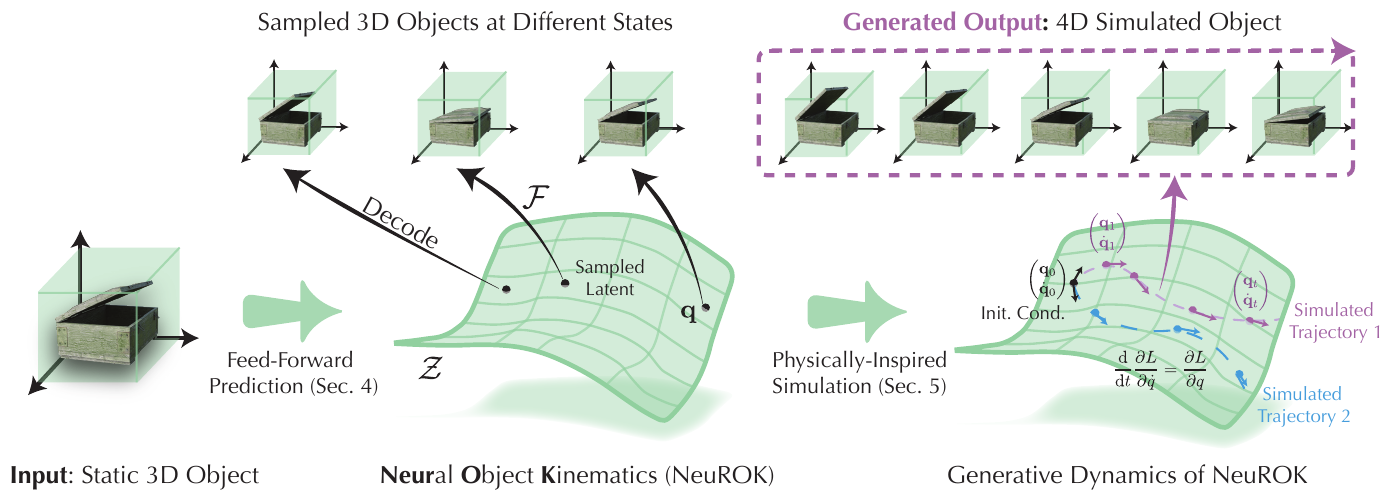}
    \vspace{-10pt}
    \caption{ \textbf{Overview of our framework.} Given a static 3D shape, \stask uses a transformer-based encoder to predict an instance-specific latent space to represent different kinematic states of this object. Each sampled latent on the learned manifold can be decoded to a corresponding state of the input object. Under different physical conditions (\textit{e.g.}, forces, actions, velocities), our method generates dynamic trajectories of latents by solving a physically-inspired ODE. 
    }
    \vspace{-14pt}
    \label{fig:overview}
\end{figure*}

\section{Overview}

\subsection{Formulation and Concepts}

This paper studies generating simulative dynamics of 3D object-centric\footnote{We colloquially define an object-centric physical system as one in which most motion arises from a single dominant deformable object.} physical systems. 
Our pipeline takes a static snapshot of a 3D dynamic object and a set of physical conditions (\textit{e.g.,} actions, forces, initial velocities) as inputs, and generates a sequence of temporally evolving 3D shapes. 
As a single 3D snapshot of an object cannot fully determine its physical parameters, our goal is to \textit{generate} one plausible 4D sequence that satisfies one valid physical  configuration and conforms to human physical intuition~\cite{battaglia2013simulation}.
We assume no kinematic or physical priors on the dynamic structure of the modeled object. The object can be articulated, rigid, a continuum body, or even a heterogeneous combination of several dynamic types, like the examples shown in \cref{fig:teaser}.

The geometry of the modeled object is represented as a mesh $\mathcal{M}_0 = (V_0, F)$ with $n$ vertices. We denote by $\xx^0 \in \rset^{3n}$ the concatenated vertex positions in $V_0$. Our pipeline outputs a sequence of deformed meshes with timestamps ranging from $1$ to $T$, denoted as $\{\mathcal{M}_1, \cdots, \mathcal{M}_T\}$, where $\mathcal{M}_t = (V_t, F)$, and the concatenated vertex positions are represented by $\xx^t \in \rset^{3n}$.

While the vertices of the mesh $\mathcal{M}_0$ can theoretically take arbitrary positions in $\rset^{3n}$, only a small subset of these configurations correspond to plausibly re-posed shapes. 
In fact, a randomly sampled deformation vector from $\rset^{3n}$ will almost certainly yield a deformed mesh far outside the distribution of valid object poses.
Empirically, the set of plausible vertex position vectors of a dynamic object forms a low-dimensional configuration manifold $\mathcal{V}^{k_\text{int}}$ embedded in $\rset^{3n}$, where $k_\text{int}$ denotes the intrinsic degrees of freedom of the deformation space and $k_\text{int} \ll 3n$. 

When studying these object-centric physical systems containing a deformable mesh with $n$ vertices, we need to define a parameterization  scheme for its kinematic states, which in turn determines the solution space for a physical simulator. We formulate this with the following definition:

\vspace{-3pt}
\begin{definition}{Kinematic State Parameterization}{kin_state_param}
A $k$-dimensional \textit{kinematic state parameterization} for a dynamic object is a pair 
$\left(\mathcal{Z}, \mathcal{F}\right)$, where $\mathcal{Z} \subseteq \mathbb{R}^k$ is the state space of the parameterization, and $\mathcal{F}: \mathcal{Z} \to \rset^{3n}$ maps any state vector $\zz \in \mathcal{Z}$ to a vertex configuration of $\mathcal{M}_0$ with $n$ vertices.
\vspace{-1pt}
\end{definition}
\vspace{-4pt}

Determining a kinematic state parameterization is the first step when studying a physical system, and it dictates how the system should be solved. As in \cref{fig:kin_state_param}(a), concise symbolic parameterizations are commonly used to simplify the solution space, but such representations are generally inaccessible in 4D generation where only the raw 3D geometry $\mathcal{M}_0$ is given. Consequently, most approaches adopt geometry-derived parameterizations, such as the high-dimensional particles (material points) used in MPM~\cite{jiang2016material}. Such parameterizations are commonly redundant and under-constrained since some configurations will yield implausibly deformed shapes, as in \cref{fig:kin_state_param}(b).

To solve dynamics in high-dimensional solution space defined by the redundant parameterization, prior works introduce category-specific physical equations and constraints to prevent the system from being under-determined. These formulations are effective in targeted domains, yet they struggle to model objects beyond the designated category.

\subsection{Proposed Solution}

\label{ssec:our_solution}

We address the above-discussed problem by introducing a kinematic state parameterization learned from data:

\vspace{-3pt}
\begin{definition}{Neural Object Kinematics}{neurok}

If a $k$-dimensional kinematic state parameterization $\left(\mathcal{Z}, \mathcal{F}\right)$ uses a neural network to represent $\mathcal{F}$, and the range of $\mathcal{F}$ is $\mathcal{V}^{k_\text{int}}$, we refer to this pair as a \btask (\stask) of a dynamic object.
\vspace{-1pt}    
\end{definition}
\vspace{-2pt}

We train an encoder-decoder model to infer \stask of a given object $\meshz$. The model comprises an encoder that encodes $\meshz$ to an instance-specific latent space $\mathcal{Z}$ of the object's kinematic states and a decoder $\mathcal{F}$ that decodes any sampled latent to a plausibly deformed shape. This model is learned with a generative objective, as detailed in \cref{sec:neurok}.

A successfully learned \stask greatly simplifies the solution space of the physical system, since we only need to model the dynamics between latent vector $\zz$ in a low-dimensional space. It also eliminates the need for inter-particle physical equations employed in mainstream simulation approaches to keep the deformed shape intact and plausible, as any sampled latent can be mapped into a validly deformed mesh. This allows us to study the system as a whole by considering the energy landscape over different kinematic states of an entire system.  

Formalizing this intuition, we simulate this system from the Lagrangian mechanics' perspective in classical physics. The learned \stask can be seen as the \emph{generalized-coordinates} of the object-centric physical system, and such systems can be solved in a generic manner by defining the Lagrangian function of the system and solving Euler-Lagrange equations~\cite{landau1960mechanics}. We detail this process in \cref{sec:simulation}.

An overview of our framework can be found in \cref{fig:overview}.

\section{Generative Learning of \stask}

\label{sec:neurok}

\begin{figure}[t]
    \centering
    \includegraphics[width=0.9\linewidth]{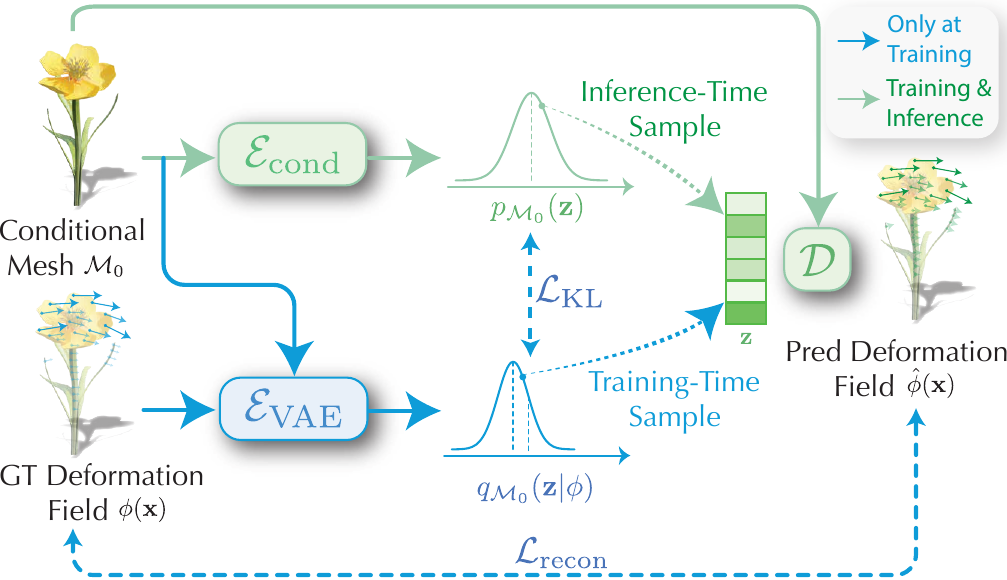}
    \vspace{-5pt}
    \caption{ \textbf{Generative learning of \stask.} During training, we randomly sample an instance mesh and one of its possible deformation fields from the training set, and supervise all three models with KL and reconstruction targets. During inference, we only use $\mathcal{E}_{\text{cond}}$ to obtain the prior distribution $p_{\meshz}(\zz)$ for the instance $\meshz$ and sample from this distribution a latent, which is further decoded to a predicted deformation field with decoder $\mathcal{D}$. 
    }
    \vspace{-15pt}
    \label{fig:neurok_arch}
\end{figure}

This section discusses the methodology of learning an encoder-decoder model to predict a \stask $(\mathcal{Z}(\mathcal{M}_0), \mathcal{F}(\cdot; \mathcal{M}_0))$ from an input mesh $\mathcal{M}_0$ of a 3D snapshot of a dynamic object. We model the latent state space $\mathcal{Z}(\mathcal{M}_0)$ associated with $\mathcal{M}_0$ by studying a surrogate task: learning a generative distribution $p_\meshz(\deformf)$ over all plausible deformation fields\footnote{To parameterize deformation fields for use in neural networks, we sample points on the mesh and treat their deformations as the parameterization of $\phi$.} $\phi(\xx): \rset^3 \to \rset^3$ of $\mathcal{M}_0$. Concretely, we train a conditional variational auto-encoder~\cite{kingma2013vae} to learn three models to approximate the instance-specific prior distribution $p_{\mathcal{M}_0}(\deformf)$:

\begin{enumerate}
    \item A \textbf{kinematic prior encoder} $\econd(\mathcal{M}_0)$ that takes in the conditioning input mesh and outputs the parameters for a prior distribution $p_\meshz(\zz)$ over the latent space $\rset^k$.
    \item A \textbf{variational deformation encoder} $\evae(\deformf, \meshz)$ that takes in a deformation field $\deformf$ and a conditional mesh $\meshz$ and produces the parameters of a posterior distribution $q_{\meshz}(\zz \mid \deformf)$. 
    \item A \textbf{deformation decoder} $\decoder(\zz, \meshz)$ that takes in a sampled latent $\zz$ from the conditional prior distribution $p_\meshz(\zz)$ and decodes it into a deformed mesh $\mathcal{M}_{\zz}$. 
\end{enumerate}

After learning these three models, we extract the high-density region of the latent probability distribution $p_\meshz(\zz)$
as the \stask kinematic state space $\mathcal{Z}(\meshz)$, and use the probabilistic decoder $\mathcal{D}(\zz, \meshz)$ as the \stask mapping $\mathcal{F}(\cdot; \meshz)$. 

An overview of these models can be found in \cref{fig:neurok_arch}. We design these three models with scalable transformer-based architectures and train them on a large-scale 4D dataset to let them learn generalizable kinematic priors.

\subsection{Model Architecture}

We now discuss the model architectures of $\mathcal{E}_{\text{cond}}, \mathcal{E}_{\text{VAE}},$ and $\mathcal{D}$. As a general principle, we use transformers~\cite{vaswani2017attention} as backbones to ensure that they scale well to large-scale datasets. 

\myparagraph{Kinematic Prior Encoder.} $\econd$ takes a conditional mesh $\meshz$ as input and outputs the kinematic prior distribution for $\meshz$. To encode $\meshz$, we evenly sample $\nsample$ points from the surface of the input mesh to form a point cloud $V_\text{sample}$.  We then use the position embedding layer following 3DShape2Vecset~\cite{zhang20233dshape2vecset} to obtain point-wise features $\pfeat \in \rset^{\nsample \times \fdim}$, where $\fdim$ is the feature dimension of position embeddings. To allow the encoder to take varying numbers of point samples from a single mesh during encoding, we adopt a perceiver-based architecture~\cite{jaegle2021perceiver,zhao2023michelangelo} and store a series of learnable tokens $\codebookcond$, where $\ntokenscond$ is the number of tokens. With the learnable tokens, we apply multiple blocks of cross-attention and self-attention layers to obtain $\ntokenscond$ encoded features $\{\mathbf{f}_i\}_{i=1}^{\ntokenscond}$. We flatten the features to form $\mathbf{\mu}_\text{cond} \in \rset^{\ntokenscond \times F_{\text{token}}}$, where $F_{\text{token}}$ is the dimension of each token. We use the normal distribution $\mathcal{N}(\mathbf{\mu}_{\text{cond}}, \mathbf{I})$ as the instance-specific prior distribution $p_\meshz(\zz)$. 

\myparagraph{Variational Deformation Encoder.} $\mathcal{E}_\text{VAE}$ outputs posterior distribution $q_\meshz(\zz \mid \deformf)$ by taking two inputs: a deformation field $\phi$ and an instance-specific mesh $\meshz$. 
To parameterize these inputs, at training time, we sample a deformed mesh of $\meshz$. This deformed mesh is represented as $\mathcal{M}_{\zz} = (V_\zz, F)$ with a shared topology $F$ as $\meshz = (V_0, F)$. Similar to $\econd$, we sample a point cloud $V_\text{sample}$ on the surface of $\meshz$. We then compute the vertex deformation\footnote{Practically, we parameterize the deformation of each point with dual quaternions. See the \supp for more discussion.} $\delta_\zz = V_\zz - V_0$ from $\meshz$ to $\mathcal{M}_\zz$ and use barycentric interpolation to compute the deformation vector $\delta_{\text{sample}} \in \rset^{d_{\text{deform}} \times \nsample}$ of the sampled points, where $d_{\text{deform}}$ is the dimensionality of the deformation representation. We then concatenate $\delta_{\text{sample}}$ with the position vectors of $V_\text{sample}$ and encode it using the position embedding layer~\cite{zhang20233dshape2vecset} to get the point-wise feature $\pfeatvae$ as inputs to the transformer. We similarly use a perceiver-based~\cite{jaegle2021perceiver} architecture and store $2\times \ntokens$ learnable tokens. These tokens are mapped to $2\times \ntokens$ features $\{\mathbf{f}_i^{\text{VAE}}\}_{i=1}^{2\times \ntokens}$. We separate those features into two sets and flatten them to represent the mean $\mu_{\text{VAE}} \in \rset^{\ntokens \times \dimtoken}$ and variance $\sigma_{\text{VAE}} \in \rset^{\ntokens \times \dimtoken}$ of the posterior. The output posterior distribution $q_\meshz(\zz \mid \deformf)$ is modeled as a Gaussian distribution $\mathcal{N}(\mu_{\text{VAE}}, \sigma_{\text{VAE}})$.

\myparagraph{Deformation Decoder.} $\mathcal{D}(\zz, \meshz)$ is a decoder that decodes sampled latent $\zz$ to a deformed mesh from $\meshz$. To implement this, we sample $\nsample$ points from the surface of the input mesh to form a query point cloud $V_{\text{query}}$. As the latent space has a dimensionality of $K \times \dimtoken$, we reshape $\zz$ into $K$ latent tokens $\{\ee_i\}_{i=1}^{K}$, each with $\dimtoken$ dimensions. We then pass the query point cloud and the latent tokens to several blocks of self-attention and cross-attention layers, and predict $\nsample$ features $\{\mathbf{f}_i\}_{i=1}^{\nsample}$. We further pass the features into an MLP to get the final deformation vectors $\delta_{\text{pred}} = \{\delta_i\}_{i=1}^{\nsample}$. We deform $V_\text{query}$ using the predicted deformation vectors, and drive the mesh vertices $V_0$ by averaging the deformations over $K_{\text{drive}}$ nearest sampled points.

\subsection{Dataset and Training}

All three models are trained simultaneously on a large-scale 4D dataset of deforming meshes of dynamic objects. We construct this  dataset by curating instances from existing works~\cite{partnet_mobility,deitke2023objaverse} and physical simulation. The details of the dataset can be found in the \supp. 

At each training iteration, we randomly sample an instance from all training instances of the dataset. For this instance, we randomly select two frames in its deformation sequence and obtain two meshes with shared topology. We use the first mesh as $\meshz$ and sample the deformation from the first mesh to the second mesh to form the sampled deformation vector $\delta_{\text{sample}}$. These are passed into three models to get the reconstructed deformation $\delta_{\text{pred}}$.
The models are supervised with the standard conditional VAE target:
\begin{equation}
   \mathcal{L} = ||\delta_{\text{sample}} - \delta_{\text{pred}}||_2^2 + \lambda D_{KL}(q_{\meshz}(\zz \mid \deformf) ||p_{\meshz}(\zz)),
\end{equation}
where $\lambda$ is a hyper-parameter and we set $\lambda=0.01$. 

\subsection{Dimension Reduction}

The raw latent space of the learned VAE can be high-dimensional. To obtain a reduced-order latent space, we further perform a dimension reduction process to compress $\mathcal{Z} \subseteq \rset^k$ to a lower-dimensional latent space $\mathcal{Q} \subseteq \rset^{k_q}$, where $k_q \ll k$. 

We perform the dimension reduction through the Active Subspace Method~\cite{constantine2014active} that reduces the dimensionality of a high-dimensional space $\mathcal{Z}$ by considering a surrogate function $\mathcal{G}(\zz) = g(A\zz + \epsilon(\zz))$, where $G: \rset^{k} \to \rset$, $A \in \rset^{k_q \times k}$, and $g: \rset^{k_q} \to \rset$. In this way, the span of the rows of $A$ identifies the directions that matter for $\mathcal{G}$~\cite{bindel2018numerical}.
We define $G$ in a way that identifies the influence of $\zz \in \mathcal{Z}$ on the predicted deformation. Therefore, we formalize $G$ as the 2-norm of $\delta_{\text{pred}}$ predicted from a set of sampled points on $\meshz$.

\section{Generative 4D Simulation}
\label{sec:simulation}

With the predicted \stask, our initial task of generating a dynamic sequence of meshes is converted into generating a series of $\{\zz_i\}_{i=1}^T$, with $\zz \in \mathcal{Z}(\meshz)$.
Note that our mapping $\mathcal{F}$ in the learned \stask will map any sampled latent $\zz$ to a plausibly deformed shape that corresponds to a valid configuration of the studied object-centric physical system. This observation motivates us to use methods from Lagrangian mechanics~\cite{landau1960mechanics} to generate such dynamics.

\subsection{Preliminaries: Lagrangian Mechanics}

Lagrangian mechanics studies a physical system by defining a set of parameters $\{\qq_1, \cdots, \qq_k\}$ that completely define the state of the system in a \textit{configuration space}. Such parameters are called \textit{generalized coordinates} of the system, and their time derivatives $\dot \qq$ are called \textit{generalized velocities.}

From this perspective, $\mathcal{Z}(\meshz)$ effectively forms a configuration space of the studied object-centric physical system, and any $\zz \in \mathcal{Z}(\meshz)$ is a vector of generalized coordinates of the system. Therefore, we can generate the dynamics of $\zz$ by using principles in Lagrangian mechanics.

Lagrangian mechanics solves the dynamics of generalized coordinates by defining a smooth function $L$ over the latent space and solving the Euler-Lagrange equation: 
\begin{equation}
   \dodt \frac{\partial L}{\partial \dot\zz} = \frac{\partial L}{\partial \zz}.
   \label{eq:euler-lag}
\end{equation}

For most physical systems we study in this paper, we define Lagrangian function $L(\zz, \zdot) = T(\zz, \zdot) - V(\zz)$ using the kinetic energy $T$ and potential energy $V$ of the system.

\subsection{Euler-Lagrange Equations for \stask}

With the defined Lagrangian functions and the learned \stask, we solve the dynamics of $\zz \in \mathcal{Z}(\meshz)$ with: 

\begin{equation}
   m G(\zz) \zdotdot + C(\zz, \zdot) + \nabla_\zz V = 0,
   \label{eq:euler-lag-z}
\end{equation}
where $G(\zz) = J_\zz^T J_\zz$, $J_\zz$ is the Jacobian of $\mathcal{F}$, $C_i = m \sum_{j, k} \Gamma_{ijk}(\zz) \zdot_j \zdot_k$, $\Gamma_{ijk}$ is the Christoffel symbol.
Its derivation can be found in the \supp. We solve it with numerical solvers and get the trajectory of $\{\zz_i\}_{i=1}^T$.

\subsection{Boundary Conditions}

Our system takes in conditions such as actions to generate 4D dynamics. They are incorporated by optimizing $(\zz_0, \dot \zz_0)$ to minimize $||\xx_0 - \mathcal{F}(\zz_0)||_2^2 + ||\dot \xx_0 - J_\zz \dot \zz_0||_2^2$,
where $\xx_0, \dot \xx_0$ are input positions and velocities of selected particles of the system. After solving $(\zz_0, \dot \zz_0)$, they serve as the initial condition for solving \cref{eq:euler-lag-z}. See the \supp for  details.

\begin{figure*}[t]
    \centering
    \vspace{-20pt}
    \includegraphics[width=0.80\linewidth]{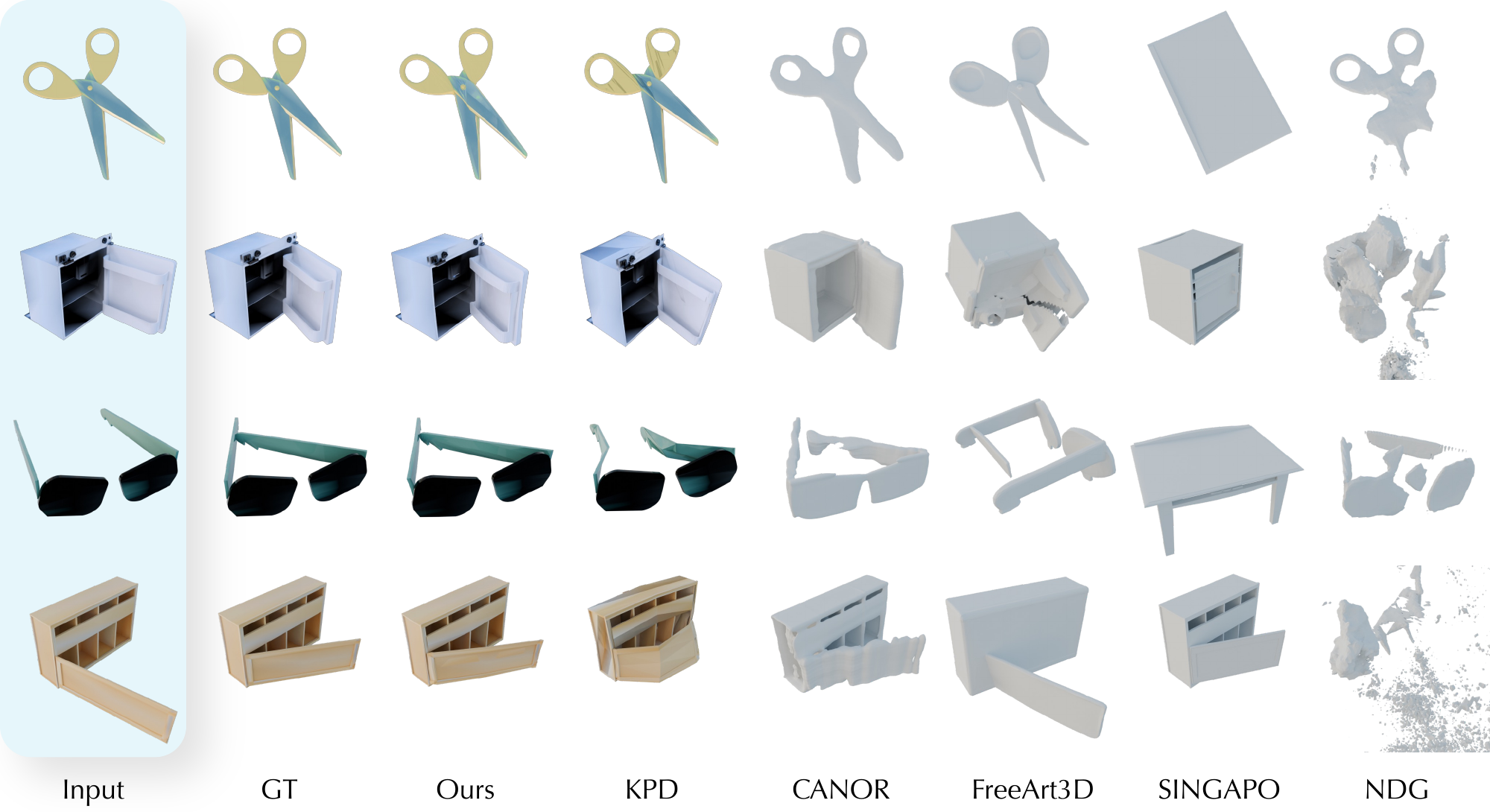}
    \vspace{-8pt}
    \caption{\textbf{Qualitative comparison on learning object kinematics.} 
    We evaluate different methods on learning compact and smooth kinematic spaces. %
    Given an input object and the shape of a target pose, we perform inverse kinematics and find the best-matching kinematic state. 
    We compare how well the reconstructed shape decoded from the obtained state vectors matches the target.
    }
    \vspace{-7pt}
    \label{fig:neurok_comp}
\end{figure*}

\begin{figure*}[t]
    \centering
    \includegraphics[width=0.85\linewidth]{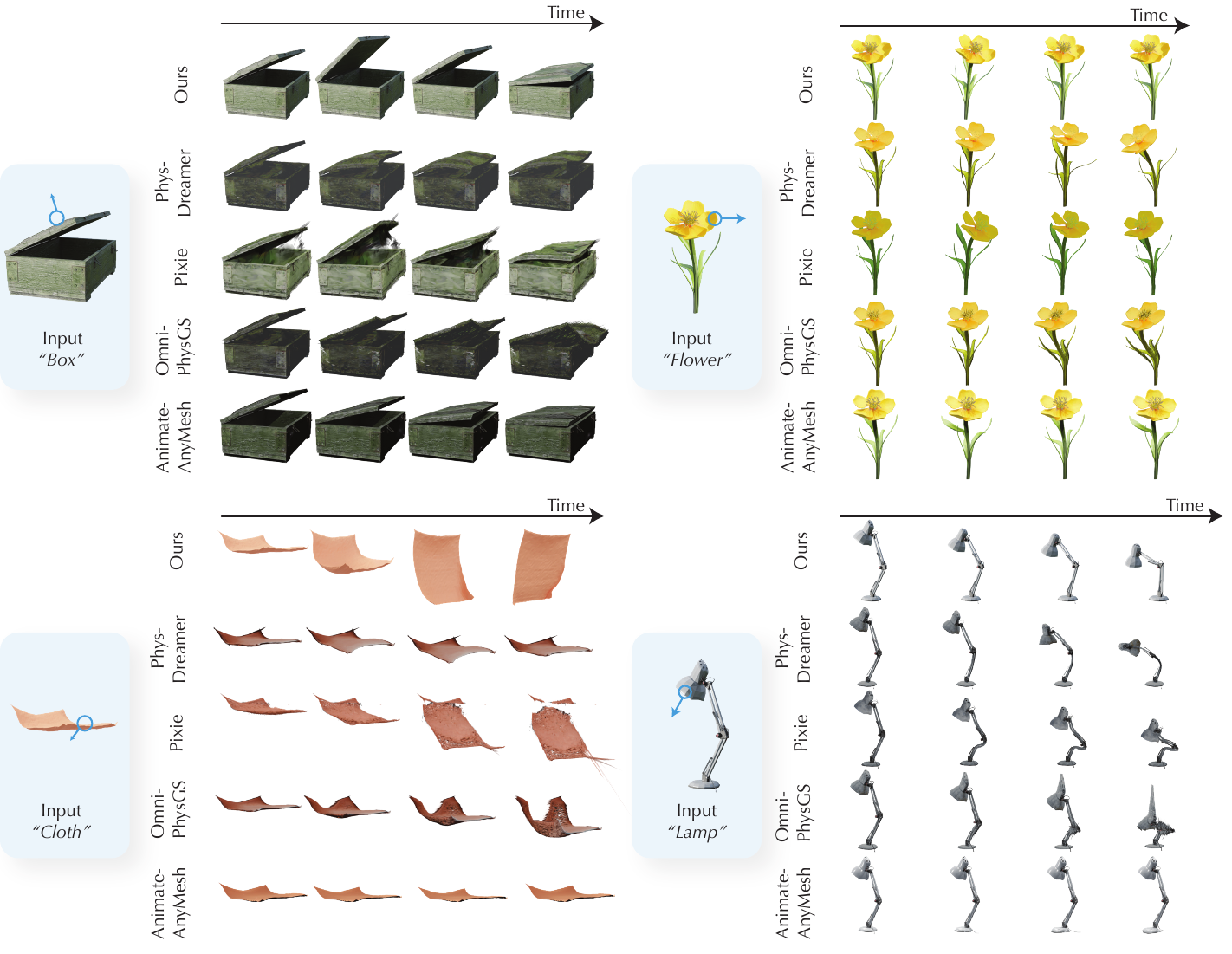}
    \vspace{-8pt}
    \caption{\textbf{Qualitative comparison on physically-inspired 4D generation.} We compare against baselines on the task of generating physically-plausible 4D motion given a single shape and conditioning actions. 
    }
    \label{fig:phy_sim_comp}
    \vspace*{-8pt}
\end{figure*}

\section{Experiments}

\subsection{\Task Learning}

We evaluate the effectiveness of \stask for learning kinematic state parameterization. %
Given an object and a target pose of the same object, we estimate an initial kinematic state and identify the optimal latent state vector that deforms the input shape to match the target.
For quantitative assessment, we use the test dataset of PartNet-Mobility~\cite{partnet_mobility}, and evaluate reconstruction accuracy using Chamfer distances~\cite{fan2017point} and a volumetric consistency metric (IoU).

\myparagraph{Baselines.} 
We evaluate two types of baselines for learning object kinematics.
NeuralDeformationGraphs (NDG)~\cite{bozic2021neural}, CANOR~\cite{he2025category}, and KeyPointDeformer (KPD)~\cite{jakab2021keypointdeformer} model kinematics using implicit representations. FreeArt3D~\cite{chen2025freeart3d} and SINGAPO~\cite{liu2024singapo} explicitly learn articulation structures, limiting them to specific object categories. %

\begin{table}[t]
    \centering
    \caption{\textbf{Quantitative comparison on inverse-kinematics optimization.}}
    \label{tab:inverse_kinematic_results}
    \vspace{-6pt}
    \resizebox{\linewidth}{!}{
    \begin{tabular}{lccc}
        \toprule
         & Chamfer (L1) \(\downarrow\) & Chamfer (L2) \(\downarrow\) & IoU \(\uparrow\) \\
        \midrule
        NeuralDeformationGraphs~\cite{bozic2021neural} & 0.670 & 0.724 & 0.289 \\
        SINGAPO~\cite{liu2024singapo} & 0.313 & 0.200 & 0.091 \\
        FreeArt3D~\cite{chen2025freeart3d} & 0.169 & 0.139 & 0.354 \\
        CANOR~\cite{he2025category} & 0.082 & 0.067 & 0.568 \\
        KeyPointDeformer~\cite{jakab2021keypointdeformer} & 0.067 & 0.067 & 0.570 \\
        \midrule
        \stask (ours) & \textbf{0.028} & \textbf{0.028} & \textbf{0.764} \\
        \midrule
        \stask w/o Model Reduction & 0.045 & 0.059 & 0.711 \\
        \stask w/o Data Augmentation & 0.036 & 0.041 & 0.724 \\
        \stask w/o Dual-Quaternion & 0.033 & 0.037 & 0.728 \\
        \bottomrule
    \end{tabular}}
    \vspace*{-7pt}
\end{table}

\myparagraph{Results.} As shown in \cref{tab:inverse_kinematic_results} and \cref{fig:neurok_comp}, our framework consistently outperforms existing methods in inverse kinematics, demonstrating its flexibility and effectiveness.

\subsection{Generative 4D Simulation}
\label{exp:gen4d}

\begin{table}[t]
    \centering
    \caption{\textbf{Quantitative comparison on physically-inspired generation.}
    We report user study preferences along with metrics from VBench~\cite{huang2024vbench} and WorldScore~\cite{duan2025worldscore}.
    AQ: Aesthetic Quality, DD: Dynamic Degrees, IQ: Imaging Quality, CLIP: CLIP score~\cite{radford2021learning}, MM: Motion Magnitude.
    }
    \label{tab:quantitative_results}
    \vspace{-6pt}
    \resizebox{\linewidth}{!}{
    \begin{tabular}{lcccccccc}
        \toprule                                               & \multicolumn{2}{c}{User Study} & \multicolumn{3}{c}{VBench~\cite{huang2024vbench}} & \multicolumn{3}{c}{WorldScore~\cite{duan2025worldscore}} \\
        \cmidrule(lr){2-3}\cmidrule(lr){4-6}\cmidrule(lr){7-9} & Alignment $\uparrow$           & Realism $\uparrow$                                & AQ $\uparrow$                 & DD $\uparrow$  & IQ $\uparrow$   & CLIP $\uparrow$ & MM $\uparrow$  \\
        \midrule {PhysDreamer~\cite{zhang2024physdreamer}}     & 5.95\%                         & 5.36\%                                            & 0.362                         & 0.500          & 48.432          & 0.716           & 0.783          \\
        {OmniPhysGS~\cite{lin2025omniphysgs}}                                   & 1.67\%                         & 0.48\%                                            & 0.380                         & 0.625          & 48.937          & 0.690           & 0.544          \\
        {Pixie~\cite{le2025pixie}}                             & 5.12\%                         & 4.17\%                                            & 0.392                         & 0.625          & 46.177          & 0.659           & 0.857          \\
        {AnimateAnyMesh~\cite{wu2025animateanymesh}}           & 5.83\%                         & 6.67\%                                            & 0.450                         & 0.625          & 48.370          & 0.730           & 0.889          \\
        \midrule \stask (ours)                                 & \textbf{81.43}\%               & \textbf{83.33}\%                                  & \textbf{0.483}                & \textbf{0.750} & \textbf{51.100} & \textbf{0.761}  & \textbf{2.343} \\
        \bottomrule
    \end{tabular}}
    \vspace*{-5pt}
\end{table}

We show that our pipeline generates 4D simulative dynamics for diverse objects, evaluated across eight objects.

\myparagraph{Baselines.} We compare against representative methods for generating 4D dynamics from 3D shapes. 
PhysDreamer~\cite{zhang2024physdreamer} distills physical parameters from video models, Pixie~\cite{le2025pixie} predicts simulation parameters using amortized-inference networks, 
and OmniPhysGS~\cite{lin2025omniphysgs} represents each asset with material-aware Constitutive Gaussians for general physics-based dynamics. 
AnimateAnyMesh~\cite{wu2025animateanymesh} is an end-to-end 4D generator trained on large-scale 4D data.

\myparagraph{Metrics.} Predicting 4D dynamics from 3D shapes is inherently ambiguous, so we evaluate plausibility and visual quality of the generated motions.
A user study with $105$ users assesses action alignment and realism. We also report metrics from VBench~\cite{huang2024vbench} and WorldScore~\cite{duan2025worldscore}.

\myparagraph{Results.} Quantitative and qualitative comparisons in \cref{tab:quantitative_results} and \cref{fig:phy_sim_comp} show that existing baselines perform well only within their specialized domains. Physically based methods (PhysDreamer~\cite{zhang2024physdreamer}, OmniPhysGS~\cite{lin2025omniphysgs}, Pixie~\cite{le2025pixie}) can handle certain material categories but generalize poorly, while end-to-end methods (AnimateAnyMesh~\cite{wu2025animateanymesh}) lack fine-grained conditioning and struggle on rarely encountered object types.
Across all settings, our method consistently generates the most physically plausible and visually realistic 4D dynamics, demonstrating strong generalization to diverse object categories.

\begin{figure}[t]
    \centering
    \includegraphics[width=0.8\linewidth]{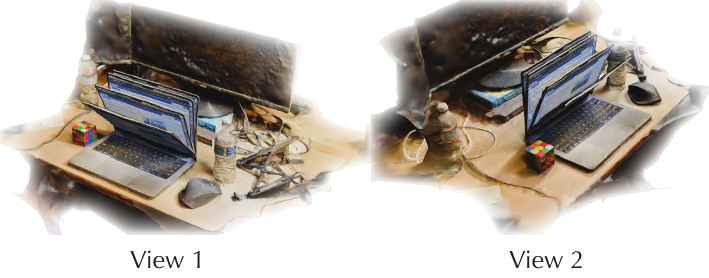}
    \vspace{-8pt}
    \caption{ \textbf{Simulating real objects.} Our model can be used to simulate real-captured objects. See the \supp for more results.}
    \label{fig:exp_real}
    \vspace*{-15pt}
\end{figure}

\myparagraph{Simulating Real Objects.} Our pipeline can also simulate and manipulate real scenes. We scan a real scene and apply our approach to simulate the dynamics of the objects within it. As shown in \cref{fig:exp_real}, our method successfully simulates the closing motion of the laptop on the desk.

\subsection{Analysis and Ablation Studies}
\vspace{-5pt}
\begin{figure}[t]
    \centering
    \includegraphics[width=0.9\linewidth]{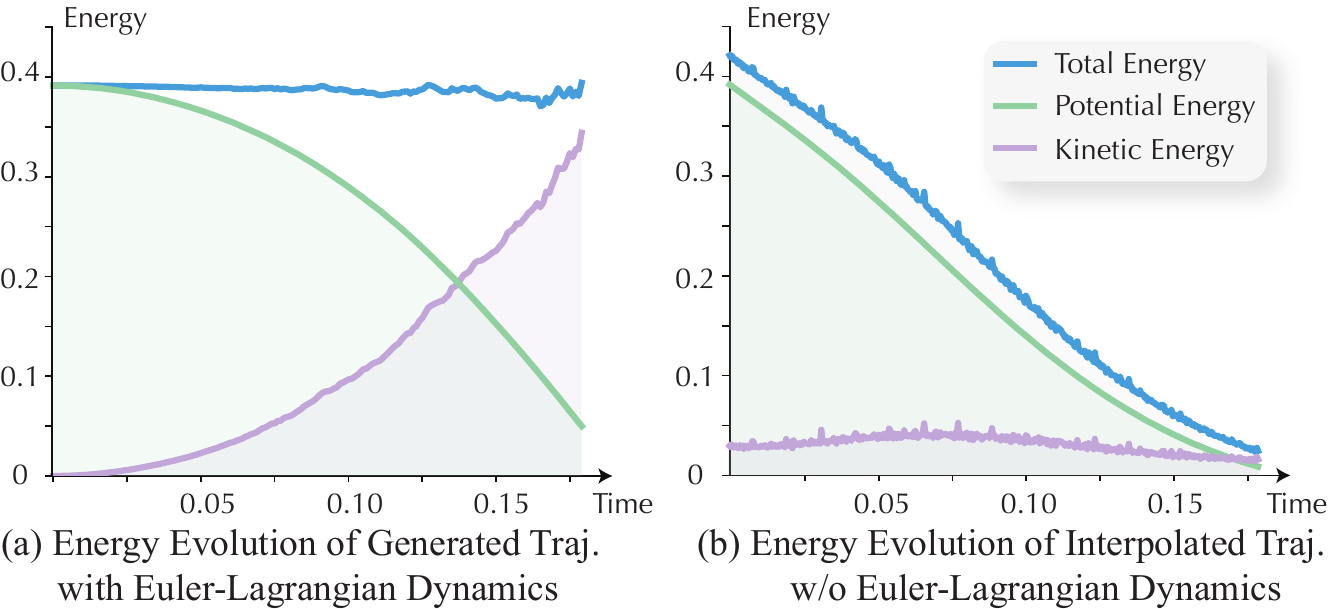}
    \vspace{-6pt}
    \caption{ \textbf{Analysis of energy conservation.} Our approach maintains physical consistency in the generated trajectories through Euler–Lagrangian modeling. Under this formulation, the total energy of the simulated motion remains approximately constant.
    }
    \label{fig:energy}
    \vspace*{-5pt}
\end{figure}

\myparagraph{Analysis of Physical Consistency.} We analyze the physical consistency of the proposed framework in ~\cref{fig:energy}. As demonstrated, our method preserves the basic conservation law of energy by leveraging a physically-inspired framework from Lagrangian mechanics.

\begin{figure}[t]
    \centering
    \includegraphics[width=0.9\linewidth]{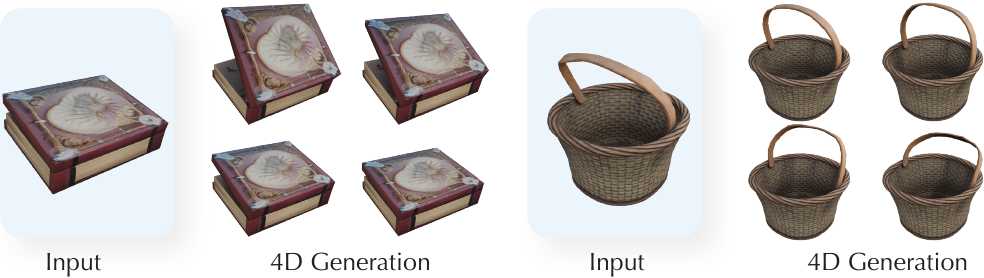}
    \vspace{-8pt}
    \caption{ 
    \textbf{Generalization on unseen categories.} Our model can generalize to novel object categories that are completely not present in the training data.
    }
    \label{fig:generalization}
    \vspace*{-14pt}
\end{figure}

\myparagraph{Generalization on Unseen Categories.} Our method learns common dynamic structures from the training dataset and successfully generalizes them to entirely new object categories. As shown in \cref{fig:generalization}, a \stask variant trained only on PartNet-Mobility~\cite{partnet_mobility} categories can still generate plausible dynamics for unseen object types.

\myparagraph{Ablation Studies.} We evaluate the impact of key design choices in \cref{tab:inverse_kinematic_results}. The results show that model reduction, training data augmentation, and our deformation parameterization each contribute significantly to the overall performance of the proposed framework.

\vspace{-4pt}
\section{Conclusion}
\vspace{-1pt}
We have introduced a novel framework, \stask, for generating 4D simulative dynamics from static 3D shapes, bridging physical principles and learned latent spaces through a physically inspired formulation. Our work opens up promising future research directions and introduces a new research paradigm in 4D visual generation.

\paragraph{Acknowledgments.} This work is in part supported by NSF RI \#2211258 and \#2338203, ONR YIP N00014-24-1-2117, ONR MURI N00014-22-1-2740, the Stanford Institute for Human-Centered AI (HAI), and the Magic Grant from the Brown Institute for Media Innovation. We acknowledge the compute support from the NSF ACCESS program \#CIS250696, Stanford Data Science and Marlowe Computing Platform, and the AMD University Program for AI \& HPC Cluster. We thank Robyn Lockwood (Stanford Language Center) for editorial and writing suggestions that improved the clarity of the manuscript. We thank Chong Zeng and Ruocheng Wang for early feedback on the manuscript and members of Stanford Vision and Learning Lab and Stanford Graphics Lab for fruitful discussion.

{
    \small
    \bibliographystyle{ieeenat_fullname}
    \bibliography{ref}
}

\end{document}